\documentclass[11pt,a4paper]{article}
\usepackage[hyperref]{emnlp2018}

\usepackage{times}
\usepackage{latexsym}
\usepackage{todonotes}
\usepackage{amsmath}
\usepackage{amssymb}
\usepackage{color,soul}
\usepackage{multirow}
\usepackage{pgfplots}
\usepackage{url}
\usepackage{xcolor}

\usepackage{graphicx}

\aclfinalcopy % Uncomment this line for the final submission

\DeclareMathOperator*{\argmax}{arg\,max}

\newcommand{\squishlist}{
 \begin{list}{--}
  { \setlength{\itemsep}{0pt}
     \setlength{\parsep}{3pt}
     \setlength{\topsep}{3pt}
     \setlength{\partopsep}{0pt}
     \setlength{\leftmargin}{1.5em}
     \setlength{\labelwidth}{1em}
     \setlength{\labelsep}{0.5em} } }

\newcommand{\squishend}{
  \end{list}  }

\usepackage{url}

\title{Improving Generalization by Incorporating Coverage \\in Natural Language Inference}

\author{Nafise Sadat Moosavi \and
Prasetya Ajie Utama \and
Andreas R\"uckl\'e \and
Iryna Gurevych\\
	Ubiquitous Knowledge Processing Lab (UKP)\\
	Department of Computer Science, Technische Universit\"{a}t Darmstadt\\
	{\url{www.ukp.tu-darmstadt.de}}\\
}

\date{}

\begin{document}
\maketitle
\begin{abstract}
The task of natural language inference (NLI) is to identify the relation between the given premise and hypothesis.  
While recent NLI models achieve very high performances on individual datasets,
they fail to generalize across similar datasets.
This indicates that they are solving NLI datasets instead of the task itself.
In order to improve generalization, we propose to extend the input representations with an abstract view of the relation between the hypothesis and the premise, i.e., how well the individual words, or word n-grams, of the hypothesis are covered by the premise.
Our experiments show that the use of this information considerably improves generalization across different NLI datasets without requiring any external knowledge or additional data.
%Finally, we show that using the coverage information is not only beneficial for improving the performance across different datasets of the same task. The resulting generalization improves the performance across datasets that belong to similar but not the same tasks.
Finally, we show that using the coverage information also improves the performance across similar tasks such as reading comprehension and QA-SRL.
%Our proposed solution  
%incorporating coverage consistently and considerably improves generalization for different NLI models and across different datasets.   
%Existing NLI approaches overly rely on lexical forms, assuming that the model learns the required abstractions.
%Instead of enriching the dataset or the model with augmented data or external knowledge to improve generalization,
%The model generalization can be improved by augmenting the dataset with more variations of the inference type
%The progress in natural language inference is mainly evaluated on large crowdsourced datasets, which contain artifacts and do not necessarily present the natural distribution of the task.
%, i.e.,
%the model that is trained on SQuAD can better answer the questions from the QA-SRL dataset. 
\end{abstract}

\section{Introduction}
%In order to facilitate the advances in various NLP applications, creating large-scale datasets using crowdsourcing has became more popular.
The task of Natural language inference (NLI) \cite{W03-0906,dagan2006pascal,D15-1075} is to specify whether the given hypothesis entails, contradicts, or is neutral regarding the premise.
While existing NLI models have high performances on individual datasets,\footnote{E.g. 91\% accuracy on the SNLI dataset \cite{liu2019multi}.} they fail to generalize across different datasets of the same task.
%As a result, the progress on such datasets, may be the result of overfitting to the dataset, and its artifacts, and not the general progress on solving the task itself.
This indicates that existing models are overfitting to specific properties of each dataset instead of learning the higher-level inference knowledge that is required to solve the task. 
%In particular, it is shown that the largest available NLI datasets, which are the basis for evaluating progress in this field, contain artifacts that are easily exploitable by neural models \cite{N18-2017}. %,S18-2023 therefore, do not well present the natural distributions of the task.

Existing solutions to improve the performance of NLI models across datasets include (1) using external knowledge \cite{joshi2018pair2vec,P18-1224}, and (2) fine-tuning on the target datasets, e.g., \cite{bansal:AAAI:2019,liu2019inoculation}.
The above approaches include the use of additional data or knowledge sources. Besides, fine-tuning on one target dataset may decrease the performance on the other datasets \cite{bansal:AAAI:2019}.

%In order to show that current models capture dataset- or word-level information, instead of high-level inference knowledge, several adversarial variations of NLI datasets are proposed, each of them reveals a specific weakness of NLI models or datasets, e.g., \cite{mccoy2019right,C18-1198,P18-2103}.
%The common approach to address the identified weaknesses is adversarial data augmentation, i.e., to integrate adversarially augmented data into the training data so that the model's performance improves on augmented weaknesses.
%The limitation of adversarial augmentation is its narrow scope. Training on one type of adversarially augmented data does not generalize to, and in fact decreases the performance on, other types of adversaries \cite{bansal:AAAI:2019}.
%Another approach to improve generalization is to incorporate background knowledge \cite{joshi2018pair2vec,P18-1224}, which requires using external resources.
In this paper, we propose a simple approach to improve generalization by using the information that is \emph{already present} in the dataset.
Existing NLI methods overly rely on the lexical form of the inputs, assuming that the model itself learns the required abstractions.
Instead, we propose to extend the existing input representations with a more abstract view of the relation between the hypothesis and premise, i.e., how well each word, or word n-gram, of the hypothesis is covered by the premise.
Our experiments show that using this information considerably and consistently improves generalization across various NLI datasets, i.e., from one to 30 points improvements.% in out-of-domain evaluations.

Finally, we show that beside improving the performance across different datasets of the same task, our proposed approach also improves the performance across datasets of similar tasks, i.e., reading comprehension on the SQuAD dataset and QA-SRL \cite{he-etal-2015-question}.

\definecolor{one}{rgb}{1.0, 0.64,0.45}
\definecolor{two}{rgb}{0.96, 0.45,0.38}
\definecolor{three}{rgb}{0.86, 0.27,0.32}
\definecolor{four}{rgb}{0.55, 0.0,0.0}

\begin{table*}[htb]
    \centering
    \footnotesize
    %\resizebox{\textwidth}{!}{%

    \begin{tabular}{|l|l| l|}
 \hline
                   P & [\textcolor{one}{{The man}}] [\textcolor{two}{\textbf{spoke}}] to [\textcolor{three}{\textbf{the lady}}] with the [\textcolor{four}{\textbf{red dress}}].  & \textbf{label}\\ \hline

         H$_1$&  [\textcolor{one}{{The man}}] [\textcolor{two}{\textbf{talked}}] to [\textcolor{three}{\textbf{the woman}}] in [\textcolor{four}{\textbf{red cloth}}]. & entail \\
         H$_2$&  [\textcolor{one}{{The man}}] did \textbf{not} [\textcolor{two}{\textbf{talk}}] to [\textcolor{three}{\textbf{the woman}}] in [\textcolor{four}{\textbf{red dress}}]. & contradict \\
         H$_3$& [\textcolor{three}{\textbf{The woman}}] [\textcolor{two}{\textbf{lectured}}] [\textcolor{one}{{the man}}]. & neutral \\ \hline
    \end{tabular}
    %}
    \caption{A sample premise and three hypotheses. Matching words, e.g., ``spoke'' and ``talked'', and bigrams, e.g., ``red dress'' and ``red cloth'', are specified by the same color.}
    \label{tab:coverage_exm}
\end{table*}

\section{Enhancing Word Representations with Coverage Information}
Assume $P$ and $H$ are the premise and the hypothesis, respectively, and $\Phi(P)\in \mathbb{R}^{|P|\times d}$ and $\Phi(H) \in \mathbb{R}^{|H|\times d}$ are their corresponding learned representations, e.g., $\Phi_i(H)$ is the representation of the $i$th word of the hypothesis.
The similarity matrix is computed as $S=\Phi(H)\Phi(P)^T$, where $S_{ij}$ shows the similarity of the $i$th word of $H$ to the $j$th word of $P$.\footnote{Such a similarity matrix is already used in various NLI models, e.g., our baseline models, for computing word alignments and attention weights.}

The \textbf{coverage values} are computed as $c_i = max_j(S_{ij})$.
$c_i$ is the maximum similarity of the $i$th word of $H$ to the premise words indicating how well the $i$th word is covered by $P$.
%indicates how well the $i$th aspect of $H$ is covered by the premise.

Moreover, we can compute the coverage values for n-grams, i.e., $\Phi_i^\prime(H)$ is the learned representation of the word n-gram of the hypothesis that starts with the word $i$ and $c^\prime_i$ is the maximum similarity of this n-gram to the n-grams of the premise.
In this work, we use bigrams and we use a CNN with the window size two for computing $\Phi^\prime$.
The use of coverage values is motivated by the success of the relevance matching model of \newcite{rueckle:AAAI:2019} for the task of answer ranking in community question answering.
%Given a query and a set of candidate answers, the task is to rank candidate answers based on their relevance to the query.
\newcite{rueckle:AAAI:2019} rank candidate answers based on on their coverage of the question bigrams.
%They consider aspects as learned representations of bigrams, which is computed using a Convolutional Neural Network with the window size two.
%In this work, we evaluate three aspect variations including: (1) unigrams,  (2) bigrams\footnote{Which are computed similar to \newcite{rueckle:AAAI:2019}.},  and (3) both\footnote{One similarity matrix for unigrams and one for bigrams.}.
Beside coverage values, we can also benefit from \textbf{coverage positions}, i.e., \mbox{$q_{i}=\argmax_j (S_{ij})$}, 
which specifies the position of the premise word (n-gram) that has the highest similarity with the $i$th word (n-gram) of $H$.

Table~\ref{tab:coverage_exm} shows a premise and three different hypotheses with their corresponding labels.
The best matching words or bigrams of the hypotheses and the premise are specified with the same colors.

For instance, in H$_2$, ``talked'' is best covered by ``spoke'' in $P$. However, the word ``not'' and the bigram ``not talked'' are not well covered by $P$.
The use of coverage values can highlight these non-covered words in H$_2$.
On the other hand, all words and bigrams of $H_3$ have high similarity (coverage) with those of the premise. However, they are not covered in the right order. The incorporation of coverage positions can help the detection of such cases, e.g., in H$_3$, the first bigram is best covered at position five of P while the last bigram is covered at position zero.

%For instance, the coverage indices for the three specified bigrams in $H_3$ are 6, 3, and 2, which inform the model that these aspects are not covered in the right order.
%using coverage index can inform the model that while ``the woman'' appears in the beginning of $H_3$, it is covered by an aspect that appears towards the end of $P$, i.e., sixth bigram.

%Following \newcite{rueckle:AAAI:2019}, we consider unigram and bigrams as aspects.

\paragraph{How to Incorporate Them?}
Assume $C$ and $C^\prime$ are the coverage vectors based on the words and bigrams of the hypothesis, respectively, e.g., $c_i \in C$ is the coverage value for the $i$th word.
Similarly, let $Q$ and $Q^\prime$ be the corresponding position vectors for $C$ and $C^\prime$.

The default setting in all of our experiments is to use  only $C$, and we consider the inclusion of the $C^\prime$, $Q$, and $Q^\prime$ vectors as hyper-parameters of the model.
The other hyper-parameter is the learned representation for computing the coverage vectors, i.e., $\Phi$, that can be the input embeddings or the output of any of the intermediate layers of the network that encode the input sentences.
%For a model that have several layers for encoding each word, the resulting representation from each of those layers can be used as $\Phi$.
We determine these two parameters based on the performance on the (in-domain) development set.

%There are several layers to encode the premise and th
%hypothesis, each of them can be used as word encodings.
%the rest of the three vectors by performing hyper-parameter optimization on the development set of the training data.
%Assume that we want to incorporate all the above four coverage vectors.
We then incorporate the coverage vectors\footnote{I.e., $C$ and those of $C^\prime$, $Q$, and $Q^\prime$ that are recognized beneficial based on the results on the development set.} by concatenating them with the learned representation of the hypothesis.
For instance, if we want to incorporate all the above vectors, the learned representation that is enhanced with the coverage information will be constructed as \mbox{$\Phi_{cov}=[\Phi(H);C;C^\prime;Q;Q^\prime] \in \mathbb{R}^{|H|\times (d+4)}$}.

The use of coverage vectors requires no change in the model architecture and the only difference is the dimensionality of the word representations from $d$ to $d+k$, where $1<k<4$, in a single layer of the network, i.e., the selected layer in which the coverage vectors are incorporated.
If the model uses the same network for encoding both $P$ and $H$, we concatenate zero values to the representation of $P$ to keep the dimensions the same for both.

%In all experiments, the coverage values (and coverage indices) are concatenated with the encoding of the hypothesis.
%and whether to use coverage indices, is made based on the performance on the development set.  
%In our preliminary experiments on incorporating coverage, we find a direct correlation between the improvements on the development set and the resulting improvements in generalization.
%For instance, ESIM summarizes the sequence of the hypothesis word encodings by using maximum and average pooling. Therefore, incorporating position information in such a model is not beneficial because it cannot be .

%The computed similarity matrix is commonly used for computing attentions and attended vectors for both inputs, e.g., question-to-passage or passage-to-question attentions \cite{bidaf,McCann2018decaNLP}.

\section{Natural Language Inference}
\label{sect:nli}
\paragraph{Datasets.} The examined datasets are:
\squishlist
\item \textbf{SNLI} \cite{D15-1075} is large crowdsourced dataset in which premises are taken from image captions.
\item \textbf{MultiNLI} \cite{N18-1101} covers ten different genres but is otherwise similar to SNLI. 
Only five genres are included in the training data. The \textit{Matched} evaluation set contains the same genres as the training data while \textit{Mismatched} set includes different genres from those of training. 
\item \textbf{SICK} \cite{sick} is also a crowdsourced dataset using sentences that describe pictures or videos. It is smaller than the first two datasets. 
%\item RTE-3 test set \cite{dagan2006pascal}: this dataset is of higher quality and hand-labeled.
%It is created so that it simulates the logical and statistical modeling of natural language reasoning.
%The reason that this dataset is not used for training recent NLI models, is its small size.
\item \textbf{\newcite{P18-2103}} is an adversarially constructed dataset, in which premises are taken from SNLI and the hypotheses differ from the premise only by one word.
Compared to SNLI, it is an easier dataset for humans. However, the performance of NLI models considerably decreases on this dataset.
\squishend
\textbf{Dataset Artifacts.} The progress in NLI is mainly determined by SNLI and MultiNLI.
These datasets are created by presenting the annotators with a premise and asking them to create three hypotheses with entailment, contradiction, and neutral labels.
This dataset creation method results in various artifacts, e.g., identifying the label by only using the hypothesis \cite{S18-2023,N18-2017}.
Similarly, the analysis of \newcite{C18-1198} suggests that models that rely on lexical overlap of premise and hypothesis and other shallow lexical cues can perform well on such datasets without performing higher-level reasoning.
Models that have high performance due to learning dataset artifacts, can be easily compromised by adversarial or more challenging examples \cite{bansal:AAAI:2019,mccoy2019right,mccoy2018non,P18-2103,C18-1198}. 
Improving the generalization of NLI models across different datasets, which presumably contain different kinds of annotation biases, is a promising direction to abstract away from dataset-specific artifacts.

\paragraph{Baseline Models.}
We incorporate coverage in two NLI models including:%\footnote{Both models are depicted in supplementary materials.}
\squishlist
\item \textbf{ESIM} \cite{P17-1152}, which is one of the top-performing systems on both SNLI and MultiNLI datasets.\footnote{E.g., its performance with ELMO embeddings and ensemble methods is 89.3 on SNLI.} We also use {ELMO} embeddings \cite{N18-1202} for this system.\footnote{ESIM and BIDAF models (with ELMO) are available at \url{https://github.com/allenai/allennlp}.}
%The structure of this network is shown in Figure~\ref{}.
%As we see, ESIM applies an average and maximum pooling layer on input encodings before the final softmax layer.

%ESIM first encodes both the premise and hypothesis using LSTM networks.
%They compute a soft alignment (attention) between the encoded words of premise and hypothesis.
%The attention weights are used for computing an attended representation for each sentence, e.g., a representation of the premise in which its relation with the hypothesis is also taken into account.
%The encoded and attended representations are fed into another network, and then a fix-size representation of each of the sentences will be computed by applying maximum and average pooling layers.
\item \textbf{MQAN }\cite{McCann2018decaNLP}, which is a sequence-to-sequence model with attention mechanisms and pointer networks \cite{vinyals2015pointer}. \newcite{McCann2018decaNLP} propose to model various NLP tasks as question answering so that we can apply a single model to various NLP tasks. 
%The architecture of MQAN is shown in Figure~\ref{}.
For the task of NLI, the question would be the hypothesis, and the context is the premise.
The main difference of this architecture compared to ESIM, or similar models, e.g., \newcite{D16-1244}, 
is that it uses the encoding of all the hypothesis words as the input to the decoder.
However, in the ESIM architecture, the encoding of individual words is not used for the final decision.
Instead, each sentence representation is summarized into a fixed vector using maximum and average pooling.
%\footnote{The architecture of these two models are included in supplementary materials.}
\squishend
\paragraph{Evaluation Metric.} Similar to previous work, we use the classification accuracy for evaluation.

\begin{table*}[htb]
\centering
%\resizebox{\columnwidth}{!}{%

\begin{tabular}{ l|l|lll}

 \hline
 & \multicolumn{1}{c|}{in-domain} & \multicolumn{3}{c}{out-of-domain} \\ \hline
 & MultiNLI & SNLI & Glockner & SICK \\ \hline
 MQAN & 72.30 & 60.91 & 41.82 & 53.95 \\
% decaNLP (81000itr.) & &60.02 & 64.31 & 42.60 & 56.57  \\ 
 %+unigram-coverage & \\
 \quad + coverage & \textbf{73.84} & \textbf{65.38} & \textbf{78.69} & \textbf{54.55}  \\
 %both coverage & 73.23 & 64.57 & 58.37 & 63.53 & 51.42 \\
 %+cove & 74.78 & 63.53 & 67.71 & \\
 %+cove(50000) & 74.60 & \\
%  \hline 
%  \hline
%  ESIM & 77.45 & 66.91 & 67.54& 65.58 & 53.20\\
%  +coverage (glove?) & \\
 \hline

 %ESIM (?) & 79.81 & 69.85 & 62.69 & 53.01\\
 ESIM (ELMO) & 80.04 & 68.70 & 60.21 & 51.37\\
   \quad + coverage & \textbf{80.38} & \textbf{70.05} & \textbf{67.47} & \textbf{52.65}\\

 \hline
\end{tabular}
%}
\caption{Impact of using coverage for improving generalization across different datasets of the same task (NLI). All models are trained on MultiNLI.}
\label{nli_generalization_mnli}
\end{table*}

\section{Improving the Cross-Dataset Performance for the Same Task}
%In this section, we evaluate the use of coverage information on MQAN and ESIM models.
We train both NLI models on the MultiNLI training data.
As the development set, we use the concatenation of matched and mismatched splits of the MultiNLI development set.
%We then select our hyper-parameters, the inclusion of $C^\prime$, $P$, and $P^\prime$ as well as the layer of the network to incorporate the coverage vectors, on this development set. 
Based on the results on this development set, the use of $C^\prime$ is useful for both models.
However, incorporating $Q$ and $Q^\prime$ only benefits MQAN.
The reason is that ESIM summarizes the sequence of hypothesis words by using maximum and average pooling and these operations cannot handle position and ordering information.
MQAN has several decoding layers on top of the hypothesis representation, and therefore, it can make sense of the coverage positions.

Table~\ref{nli_generalization_mnli} shows the performance for both systems for in-domain (the MultiNLI development set) as well as out-of-domain evaluations on SNLI, Glockner, and SICK datasets.
%\footnote{We train each model multiple times.  For ``+coverage'' experiments, we observe a very low variance between different runs. For baselines, we report the evaluation of the model with the best development results.}

\begin{table*}[!htb]
\centering
%\resizebox{\columnwidth}{!}{%

\begin{tabular}{ l|rr|rr }
 & \multicolumn{2}{c|}{in-domain} &  \multicolumn{2}{c}{out-of-domain}  \\ 
 & \multicolumn{2}{c|}{SQuAD}  & \multicolumn{2}{c}{QA-SRL}  \\ 
 \hline
 & EM & F1 & EM & F1  \\ \hline
 MQAN & 31.76& 75.37 & \textbf{10.99} & 50.10 \\
 \quad +coverage & \textbf{32.67} & \textbf{76.83} & 10.63 & \textbf{50.89}  \\ \hline
 BIDAF (ELMO)& 70.43 & 79.76 & 28.35 & 49.98 \\
 \quad +coverage & \textbf{71.07} & \textbf{80.15} & \textbf{30.58} & \textbf{52.43}  \\
\hline
\end{tabular}
%}
\caption{Impact of using coverage for improving generalization across the datasets of similar tasks. Both models are trained on the SQuAD training data.}
\label{tab:qa_results}
\end{table*}

The results show that coverage information considerably improves the generalization of both examined models across various NLI datasets.
The resulting cross-dataset improvements on the SNLI and Glockner datasets are larger than those on the SICK dataset.
The reason is that the dataset creation process and therefore, the task formulation is similar in SNLI and MultiNLI, but they are different from SICK.
In particular, in the neutral pairs in SNLI and MultiNLI, the hypothesis is mostly irrelevant to the premise, e.g., ``He watched the river flow'' and ``The river levels were rising''.
However, in the SICK dataset, neutral pairs also have high lexical similarity, e.g., ``A woman is taking eggs out of a bowl'', and ``A woman is cracking eggs into a bowl''. 
As a result, a model that is trained on SNLI or MultiNLI does not learn to properly recognize the neutral pairs in the SICK dataset.

\section{Improving the Cross-Dataset Performance for Similar Tasks}
In this section, we examine whether we can benefit from the coverage information (1) in a task other than NLI, and (2) for improving performance across datasets that belong to similar but different tasks. 
%of question answering by incorporating how well each aspect of the question is covered by the given passage.
To do so, we select two related question answering tasks including reading comprehension and Question-Answer driven Semantic Role Labeling (QA-SRL) \cite{he-etal-2015-question}.
We train the baseline question answering models on a reading comprehension dataset and evaluate their performance on QA-SRL.

\paragraph{Datasets.} For training, we use the SQuAD dataset \cite{D16-1264}, in which a question and a passage are given and the task is to extract the answer from the passage.
%The SQuAD dataset is also created by crowdsourcing and contains known artifacts \cite{D18-1546}.

%In order to evaluate how well the model captures task-level, instead of dataset-level, question answering skills, we use the question-answer driven semantic role labeling (QA-SRL) dataset \cite{D15-1076} for evaluation.
%, and zero-shot relation extraction (ZRE) \cite{K17-1034}.
In QA-SRL, predicate-argument structures\footnote{I.e., ``who'' did ``what'' to ``whome''.} are present using natural question-answer pairs. For instance, in the sentence ``John published a comic book'', the predicate-argument structures for the verb ``published'' can be determined by answering the questions ``who published something?'' and ``what was published?''.
In this dataset, the wh-questions and  a sentence
%, from which such question-answer pairs are extracted,
are given and the task is to find the answer span from the sentence.

We hypothesize that if a model learns more abstract knowledge about answering questions from the SQuAD dataset, it can also perform better in answering the questions of QA-SRL.\footnote{This setting is also used for evaluating general linguistic intelligence by \newcite{yogatama2019learning}.}

%\newcite{K17-1034} model relation extraction as a reading comprehension task in which each relation slot is associated with one or more questions.
%For instance, the relation $spouse(John,y)$ can be defined by natural questions like ``who did John married?'' or ``who is John's spouse?''.
%The test data in ZRE contains new relation types that have no labeled training examples.

%While there are differences in the format of question in SQuAD, QA-SRL, and ZRE,
%if a model learns more abstract knowledge about answering questions from the given passage, it would perform better in of all these datasets.\footnote{This setting is also used for evaluating general linguistic intelligence by \newcite{yogatama2019learning}} 
%We use the standard evaluation metrics for each of the above tasks, i.e., exact match and F$_1$ in reading comprehension and QA-SRL, and corpus-F$_1$ in ZRE.  

\paragraph{Baselines.} We use BIDAF \cite{bidaf} with ELMO embeddings as well as MQAN as baselines.
Coverage vectors are computed based on the maximum similarity of the question words to those of the given context, i.e., a passage in SQuAD and a sentence in QA-SRL.
Based on the results on the SQuAD development set, incorporating $C^\prime$ coverage vector is useful for MQAN and BIDAF only benefits from using $C$.

%and incorporating only unigram coverage values in BIDAF are the best settings for QA experiments.
\paragraph{Results.}

Table~\ref{tab:qa_results} shows the impact of coverage for improving generalization across these two datasets that belong to the two similar tasks of reading comprehension and QA-SRL.
The models are evaluated using Exact Match (EM) and F$_1$ measures, which are the common metrics in QA.
The results are reported on the SQuAD development set and the QA-SRL test set. 
All models are trained on the SQuAD training data.
As the results show, incorporating coverage improves the model's performance in the in-domain evaluation as well as the out-of-domain evaluation in QA-SRL.
This indicates that more generalizable systems also improve the performance across related tasks.

\section{Conclusions}
Despite the great progress in individual NLP datasets, current models do not generalize well across similar datasets, indicating that we are solving datasets instead of tasks.
Existing NLP methods mainly rely on the lexical form of the inputs, assuming that the required abstract knowledge is learned implicitly by the model.
In this paper, we propose a simple method that requires no additional data or external knowledge, to improve generalization.
We propose to extend the input encodings with a higher-level information regarding the relation of the input pairs, e.g., the hypothesis and the premise in NLI.
We show that the proposed solution considerably improves the performance (1) across datasets that represent the same task, and (2) across different datasets that represent similar tasks.
%For this purpose, we use coverage information, i.e., how well the hypothesis (query) aspects are covered by the premise (passage).
%providing the models with a higher level representation regarding the relation of input pairs in the existing datasets.
%We show the effectiveness of enriching inputs with coverage information in various NLI and QA models and datasets.

\section*{Acknowledgements}
The authors would like to thank G\"ozde G\"ul Sahin, Yang Gao, and Gisela Vallejo for their insightful comments on the paper. 
This work has been supported by the German Research Foundation (DFG) as part of the
QA-EduInf project (grant GU 798/18-1 and grant RI 803/12-1), and by the DFG-funded research training group ``Adaptive Preparation of Information form Heterogeneous Sources'' (AIPHES, GRK 1994/1).

\bibliography{paper}
\bibliographystyle{acl_natbib_nourl}
\appendix

\end{document}